\documentclass{article}

    \usepackage[final]{neurips_2024}

\usepackage[utf8]{inputenc} 
\usepackage[T1]{fontenc}    
\usepackage{hyperref}       
\usepackage{url}            
\usepackage{booktabs}       
\usepackage{amsfonts}       
\usepackage{nicefrac}       
\usepackage{microtype}      
\usepackage{xcolor}         
\usepackage{multirow}

\usepackage{graphicx}

\usepackage{graphicx}
\usepackage{subcaption}
\usepackage{wrapfig}

\usepackage{adjustbox}
\usepackage{soul}

\title{Sparse Upcycling: Inference Inefficient Finetuning}

\author{%
  Sasha Doubov \\
  Databricks \\
  \texttt{sasha.doubov@databricks.com} \\
    \AND
  Nikhil Sardana \\
  Databricks \\
  \texttt{nikhil@databricks.com} \\
  \And
  Vitaliy Chiley \\
  Databricks \\
  \texttt{vitaliy.chiley@databricks.com} 
}

\begin{document}

\maketitle

\begin{abstract}
    Small, highly trained, open-source large language models are widely used due to their inference efficiency, but further improving their quality remains a challenge. Sparse upcycling is a promising approach that transforms a pretrained dense model into a Mixture-of-Experts (MoE) architecture, increasing the model's parameter count and quality. In this work, we compare the effectiveness of sparse upcycling against continued pretraining (CPT) across different model sizes, compute budgets, and pretraining durations. Our experiments show that sparse upcycling can achieve better quality, with improvements of over 20\% relative to CPT in certain scenarios. However, this comes with a significant inference cost, leading to 40\% slowdowns in high-demand inference settings for larger models. Our findings highlight the trade-off between model quality and inference efficiency, offering insights for practitioners seeking to balance model quality and deployment constraints.
\end{abstract}

\section{Introduction}

Recent advancements have made several small, open-source large language models (LLMs) widely available to practitioners, such as Llama 3 8B \citep{dubey2024llama}, Gemma 2B, and Gemma 9B \citep{team2024gemma}.
These dense decoder-only models are designed for inference efficiency: smaller than their flagship counterparts (often 70B+ parameters), but trained on trillions of tokens for improved quality.
Despite their high quality, there remains a persistent demand for further improvements, particularly in downstream task performance.

One common approach to improve model quality is continued pretraining (CPT), where the model is further trained on additional data, typically on a different dataset than the original pretraining data \citep{gupta2023continualpretraininglargelanguage}.
While CPT can improve model quality, it is limited by the original model’s dense architecture and parameter count.

An alternative and increasingly popular approach is sparse upcycling \citep{komatsuzaki2023sparseupcyclingtrainingmixtureofexperts}, which increases a model’s parameter count by converting a dense model into a Mixture-of-Experts (MoE) model.
MoE architectures dynamically activate only a subset of weights (experts), allowing the model to expand its parameter count without proportionally scaling up training and inference FLOPs \citep{shazeer2017outrageouslylargeneuralnetworks}.
Sparse upcycling a dense model into an MoE architecture has the potential to improve model quality, but it introduces a trade-off: sparse upcycled models are significantly larger than their dense counterparts, resulting in higher inference costs and limiting their utility for large-scale real-world deployments.
Sparse upcycling thus conflicts with the growing trend to deploy smaller models specifically to decrease the cost of inference \citep{sardana2024chinchillaoptimalaccountinginferencelanguage}.

In this work, we ask: \textbf{What is the trade-off between model quality and inference efficiency for sparse upcycling?}
We compare sparse upcycling to dense CPT across varying model sizes, compute budgets, and pretraining durations.
By exploring these trade-offs, we provide insights for practitioners on how to balance performance gains with the practical costs of deploying LLMs.

\section{Related Work}

Mixture-of-Experts (MoE) \citep{shazeer2017outrageouslylargeneuralnetworks, fedus2022switchtransformersscalingtrillion, gale2022megablocksefficientsparsetraining} is a sparse model architecture in which input tokens are dynamically routed to different weights (experts) in a model. The typical MoE language model architecture modifies the transformer block: the attention weights stay the same, but rather than having a single multi-layer perceptron (MLP) process tokens, a routing function is introduced which dynamically routes tokens to a subset of different MLPs. This form of dynamic sparsity leads to more compute-efficient training and better inference performance. Popular open-source MoEs have been released \citep{Databricks_2024, jiang2024mixtralexperts}, typically at larger model sizes than the smallest dense models available. 

Sparse upcycling was introduced in \citet{komatsuzaki2023sparseupcyclingtrainingmixtureofexperts}, which experimented with upcycling for vision and language modeling tasks. The authors duplicated a transformer's MLPs $n$ times to create $n$ MoE experts and replaced half of the MLP layers with MoE layers. They ablated many design choices, such as different routing mechanisms, pretraining durations, and adding Gaussian noise to the expert weights. Unlike our setting, the authors studied encoder-decoder T5 models \citep{raffel2023exploringlimitstransferlearning}, rather than the decoder-only auto-regressive models that are currently popular in the open-source community. \citet{komatsuzaki2023sparseupcyclingtrainingmixtureofexperts} also performed upcycling and CPT in a similar optimization setting as the pretraining phase: they use the \textit{same} data as during pretraining, and the learning rate schedule is seamlessly continued due to the use of inverse-sqrt learning rate schedule. In practice, open-source models have been cosine annealed and the original data is not available \citep{dubey2024llama}, making it difficult to apply existing results.

More recent works have more complex training recipes, by training $n$ experts on $n$ different datasets, later merging them into an MoE. \cite{sukhbaatar2024branchtrainmixmixingexpertllms} show that performing data-specific dense finetunes, averaging the attention weights, and creating different MLP experts is a more efficient training method than simply duplicating MLPs during upcycling. \cite{zhang2024bamjustlikethat} replaces the averaging step for attention with parallel attention modules. However, \cite{wei2024skyworkmoedeepdivetraining} shows that the gap between data-specific upcycling and vanilla (duplicating MLPs) sparse upcycling diminishes over longer training durations. We chose to focus on the vanilla sparse upcycling recipe in this work with further discussion in Section \ref{sec:limits}. Past work has not benchmarked inference performance of upcycled models, a focus of our work.


\section{Upcycling Quality Improvements}

\subsection{Experimental Setup}

We have two distinct phases of training: dense pretraining and continued pretraining/upcycling.
During the dense pretraining phase, we train a model on a generic common crawl, while CPT/upcycling is done with a higher quality domain data. Our dense pretrained models are fully cosine annealed, and we perform warmup during the CPT/upcycling training phase.

We experiment with two dense model sizes: a 436M and 1.4B model. Further experimental details are described in Appendix \ref{appendix:experimental}. 

\subsubsection{Pretraining}

\begin{table}[b]
  \vspace{-5mm}
  \caption{Pretraining models and training durations.}
  \label{tab:pretraining-tokens}
  \centering
    \begin{tabular}{llr}
      \toprule
      \textbf{Model Size} & \textbf{Duration} & \textbf{Tokens} \\
      \midrule
      \multirow{3}{*}{436M} & Medium      & 43B \\
                            & Long        & 100B \\
                            & Extra Long  & 200B \\
      \cmidrule(r){1-3}
      \multirow{2}{*}{1.4B} & Medium      & 142B \\
                            & Long        & 354B \\
      \bottomrule
    \end{tabular}
\end{table}

We vary the amount of pretraining tokens for the different model sizes, as shown in Table \ref{tab:pretraining-tokens}. 
Note that we train our models for a large amount of pretraining tokens relative to the model size, with further discussion in Appendix \ref{appendix:pretraining-duration}.

\subsubsection{CPT/Upcycling}

We create a Mixture-of-Experts from a dense checkpoint by duplicating the GLU layers within a block 8 times, with 8 experts and randomly initialize the router weights.
We use top-$K$ learned dropless routing following \cite{gale2022megablocksefficientsparsetraining} with $K=2$. The resulting upcycled models contain 1.6B parameters (originally 436M) and 6.7B parameters (originally 1.4B). 
We perform different CPT and upcycling runs and match FLOP budgets for each run. 

We report downstream results on the Eval Gauntlet v0.3 \citep{mosaicml2023mpt}, a benchmark of 35 in-context learning tasks. We report the aggregate Gauntlet Core Average score, which is normalized and averaged across all of the tasks. More details are available in Appendix \ref{appendix:eval}.

\subsection{Results}

    \begin{figure}[t]
        \subfloat{%
            \includegraphics[width=.48\linewidth]{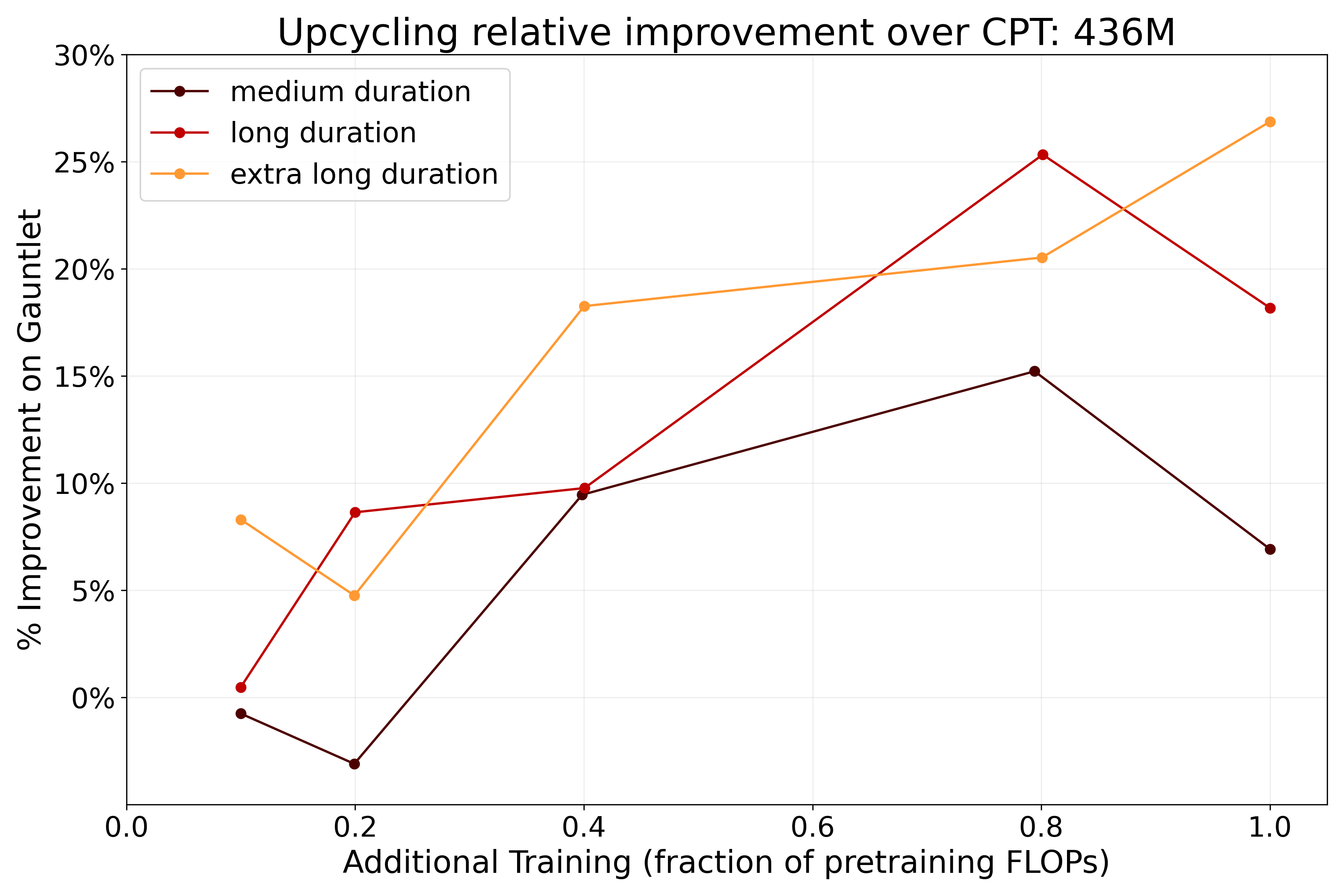}%
            \label{subfig:a}%
        }\hfill
        \subfloat{%
            \includegraphics[width=.48\linewidth]{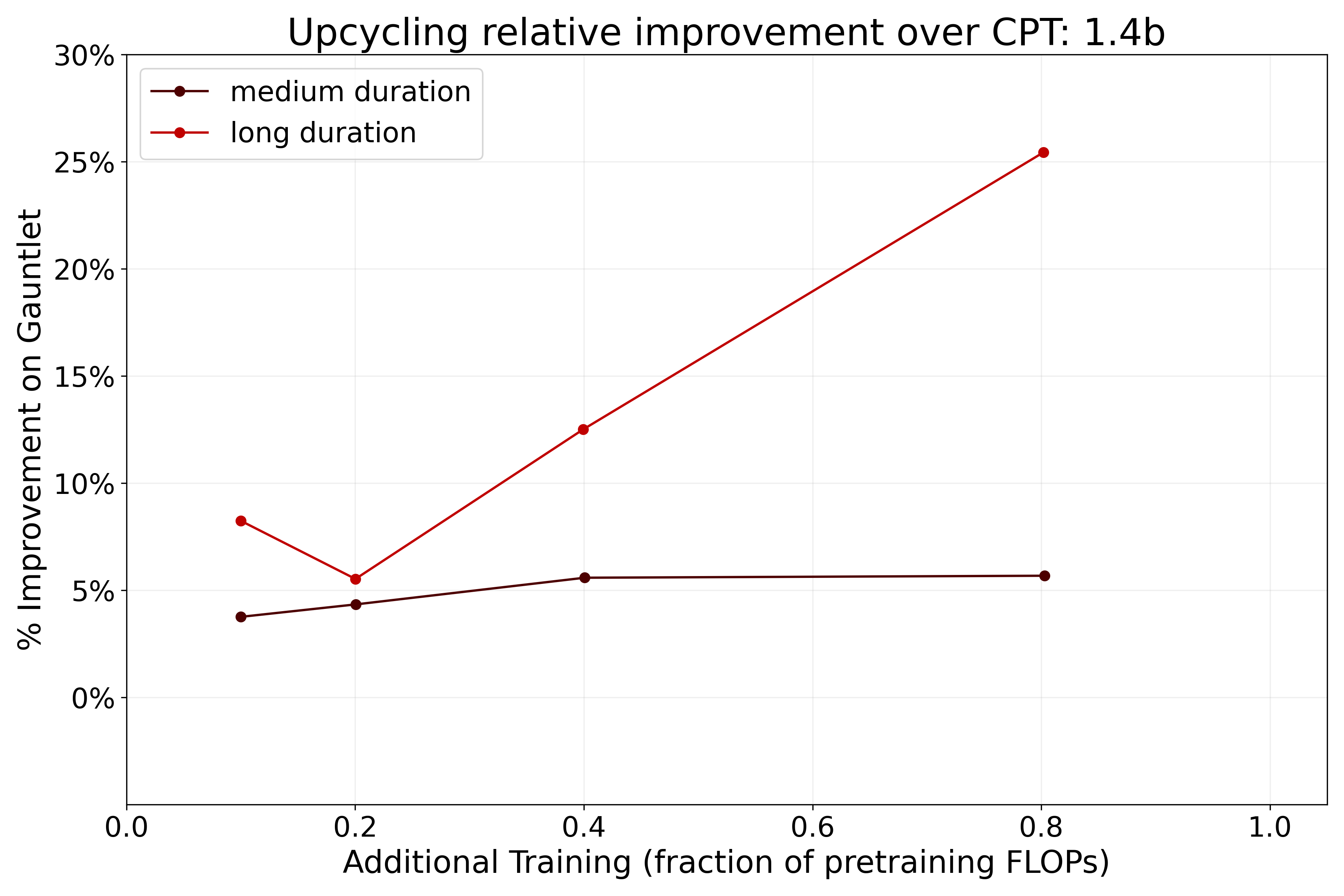}%
            \label{subfig:b}%
        }\\
        \subfloat{%
            \includegraphics[width=.48\linewidth]{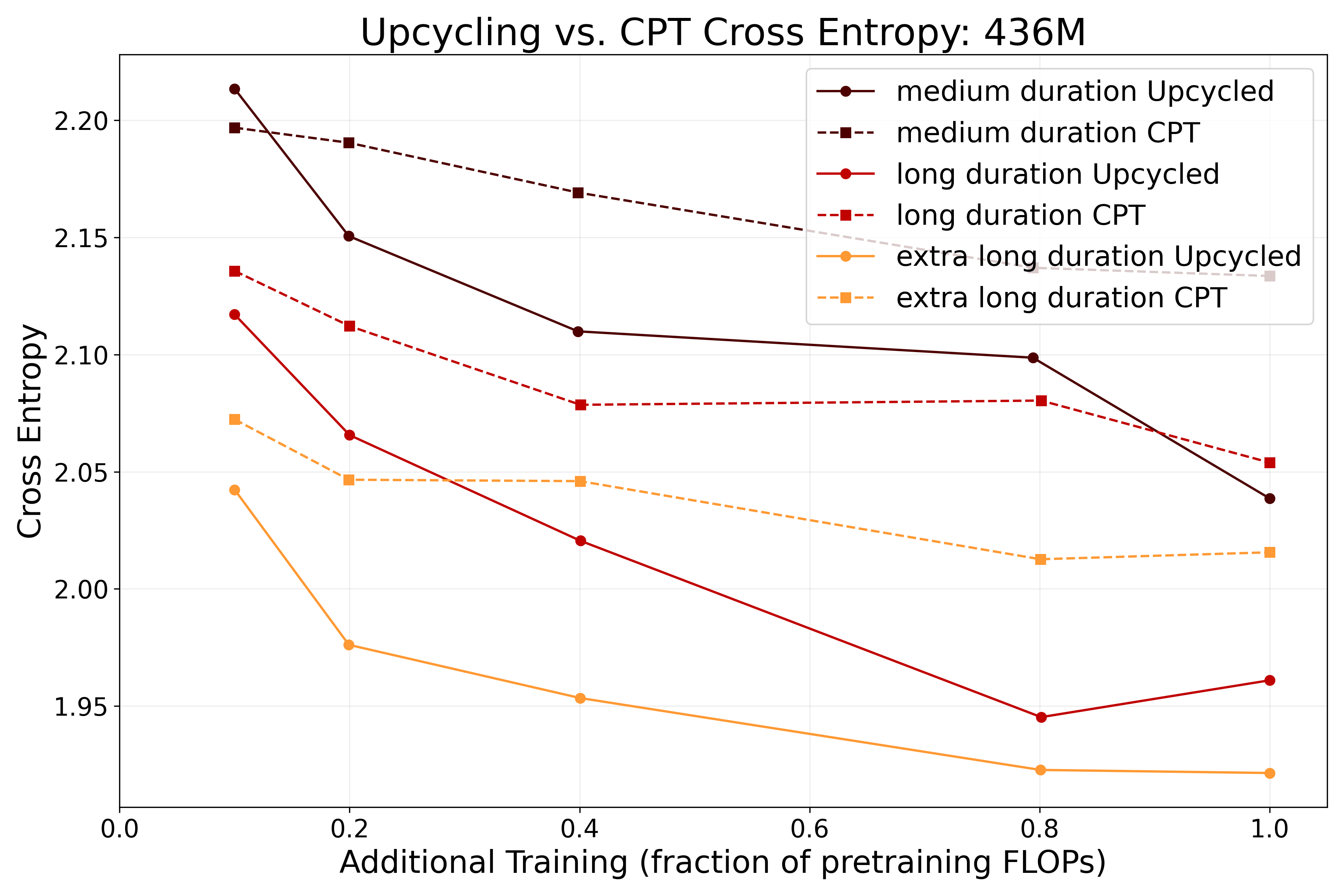}%
            \label{subfig:c}%
        }\hfill
        \subfloat{%
            \includegraphics[width=.48\linewidth]{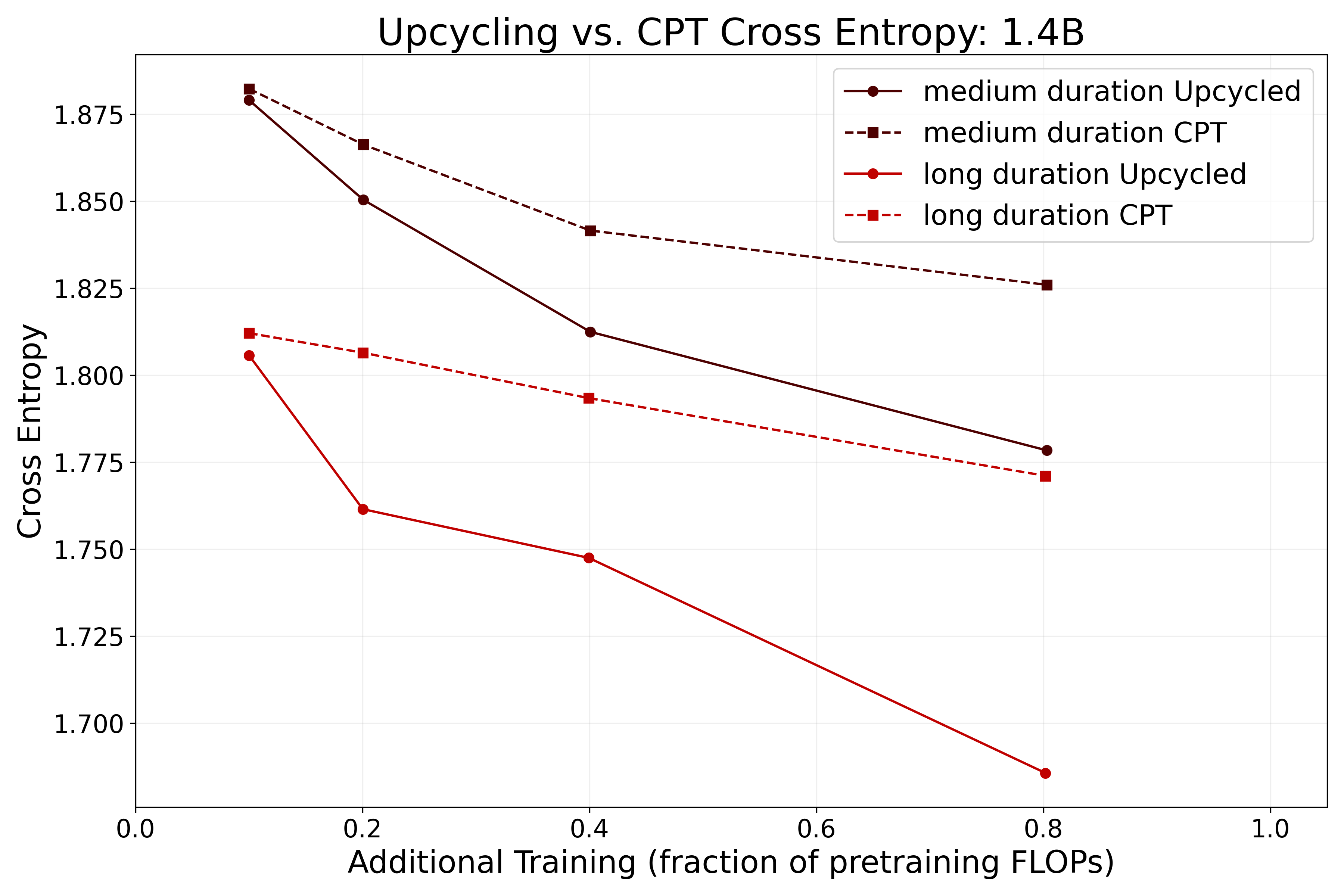}%
            \label{subfig:d}%
        }
        \caption{\textbf{Results from CPT vs. upcycling.} The top row shows the relative improvements on Gauntlet Core Average as a function of additional training. The bottom row compares the final cross entropy scores of the upcycled and CPT models.}
        \label{fig:main-train-results}
    \end{figure}

We show iso-FLOP comparisons of upcycling and CPT in Figure \ref{fig:main-train-results}. In general, upcycling tends to achieve lower loss compared to the corresponding CPT run, showing that upcycling is indeed able to produce better quality models. For both CPT and upcycling, there are diminishing returns from extending training arbitrarily, although CPT saturates earlier than the upcycling runs.

Figure \ref{fig:main-train-results} also shows the relative improvements on Gauntlet Core Average across different models and durations. We find that a significant fraction of pretraining is needed to improve downstream performance --- at 40\% of the full training budget we see gains up to 20\% with the 436M model and less than 15\% with the 1.4B model. 

Hence, upcycling is able to achieve lower loss and improve downstream quality, albeit it requires a significant portion of the original compute budget.

\section{Inference Efficiency}
\subsection{Experimental Setup}

To show the difference in inference efficiency between CPT vs. upcycled models, we benchmark latency and throughput of our 436M and 1.4B dense models along with their upcycled counterparts. In addition, we measure an 8B dense and corresponding 47B upcycled model to understand inference for upcycling a model at the popular 7--8B scale. 

For the MoE models, we include benchmarking with top-$K$=1 and top-$K$=2. This value sets the sparsity of the MoE model: a token is routed to $K$ experts as part of the FFN computation. A higher top-$K$ value leads to more active parameters and correspondingly more FLOPs used during a model pass. When an upcycled model uses top-$K$=1, it approximately matches the FLOPs of the original dense model, excluding the small overhead of the router computation.

We benchmark a typical inference workload using Nvidia's H100 GPUs and the vLLM inference engine. We measure inference performance by measuring the latency of requests and throughput across the requests. Further benchmarking details are available in Appendix \ref{appendix:inference}.

\subsection{Results}

\begin{figure}[t]
    \hspace*{-14mm}  
    \centering
    \begin{subfigure}[c]{0.45\textwidth}
        \centering
        \adjustbox{valign=m}{%
            \includegraphics[height=4cm, trim={0cm 0cm 0cm 0cm}, clip]{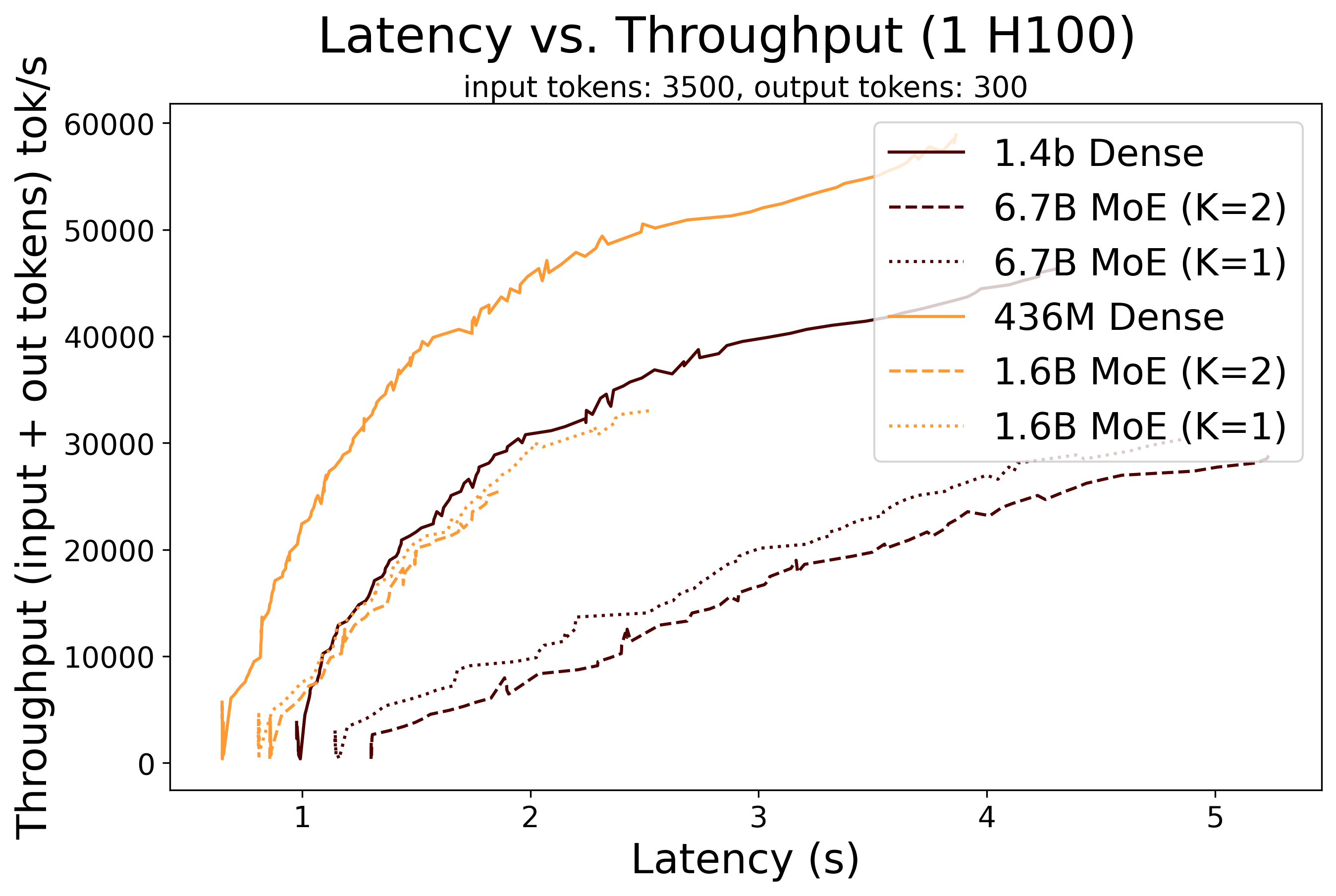}
        }
    \end{subfigure}
    \hspace{0pt}
    \begin{subfigure}[c]{0.45\textwidth}
        \centering

        \adjustbox{valign=m}{%
        \small
        \setlength{\tabcolsep}{2pt}
        \renewcommand{\arraystretch}{0.9}
        \begin{tabular}{lcccc}
            \toprule
            \textbf{Model Size} & \textbf{\# H100} & \textbf{Top-K} & \textbf{\shortstack{Max \\ Throughput}} & \textbf{\% Decrease} \\
            \midrule
            436M        & 1 & -- & 59,128 & --    \\
            1.6B (MoE)  & 1 & 1  & 37,164 & 37\%  \\
            1.6B (MoE)  & 1 & 2  & 32,832 & 44\%  \\
            \midrule
            1.4B        & 1 & -- & 48,032 & --    \\
            6.7B (MoE)  & 1 & 1  & 31,464 & 34\%  \\
            6.7B (MoE)  & 1 & 2  & 29,716 & 38\%  \\
            \midrule
            8B          & 4 & -- & 33,516 & --    \\
            47B (MoE)   & 4 & 1  & 21,356 & 36\%  \\
            47B (MoE)   & 4 & 2  & 18,924 & 44\%  \\
            \bottomrule
        \end{tabular}
    }
    \end{subfigure}
    \caption{Inference Speed of CPT vs. Upcycled Models.}
    \label{fig:upcycling-benchmarking}
\end{figure}

Dense models outperform their upcycled counterparts significantly at inference time, as shown in Figure \ref{fig:upcycling-benchmarking}. In the high request regime, we compare the maximum throughput achieved by the dense models vs. upcycled models, and find significant decreases across the board. This is especially true for the larger 1.4B and 8B models, and even holds when top-$K$=1.

In the high request regime, ie. when comparing max throughput, one would expect the performance for the dense model and the top-$K$=1 to be comparable. This is because the model is operating in the compute-bound regime, where the number of active parameters determines inference speed. However, we still see a large gap in maximum throughput between dense and MoE models. This may be due a few different factors. Dense models have fewer total parameters than their upcycled counterparts, which means that they use less GPU memory and can support higher batch sizes, allowing them to achieve higher throughput. In addition, there is likely room for MoE-specific optimizations in vLLM that can further close the gap between dense and MoE model performance \citep{moemodelpytorch}.

As mentioned above, the increased parameter count in upcycled models leads them to using more GPU memory for storing the weights. The higher memory usage limits the hardware configurations suitable for running the upcycled models or necessitates model compression techniques.

Hence, our empirical results show a large gap in MoE and dense inference performance, even in regimes where it would be expected to have less of a performance gap.

\pagebreak

\section{Discussion}

Larger upcycled models are slower for inference, with maximum throughput reductions of 35-45\% empirically. Using $K=1$ leads to smaller throughput drops, although the gap remains significant. 

We find that upcycling benefits from longer training durations, offering significant benefits over CPT after training for over 20-40\% of the original pretraining budget. Hence, a significant amount of FLOPs must be sunk into upcycling to reap the benefits of the additional parameters. Upcycling is a good fit with a large training FLOP budget and for applications where quality significantly outweighs inference performance.

Past work has not focused on the inference overhead of upcycling, emphasizing model quality. They also did not show real-world inference performance of upcycled models. However, given the growing trend of training smaller models for inference and inference-time compute \citep{snell2024scalingllmtesttimecompute}, we believe that the inference cost of upcycling is an important consideration when trying to improve model quality.

\section{Limitations}\label{sec:limits}

We note that our work has several limitations, in particular:

\begin{enumerate}
    \item \textbf{Fixed MoE architecture} We do not ablate the number of experts or top-$K$ value during training. We believe that 8 experts and $K=2$ is a reasonable architecture used in open-source models such as Mixtral \citep{jiang2024mixtralexperts} and keeps the total parameters of the model relatively low. We benchmark top-$K$=1 to see how matching MoE and dense active parameters affects real-world inference performance.

    \item \textbf{Inference Setup: } Our benchmarking setup uses vLLM as our inference engine. While this is a popular inference engine, we expect performance can differ significantly across different engines and that performance for MoEs can differ as well \citep{sglangllama3}. Still, we feel our work provides an idea of out-of-the-box performance of serving upcycled MoEs.
    
    \item{\textbf{Vanilla upcycling recipe:} We replicate the GLU layers rather than doing dense fine-tuning on different domains and then merging the models. While a more complex recipe may increase training efficiency gains, we choose to focus on the simpler setting and believe this is a promising technique for future work.}

    \item{\textbf{Assumes pre-existing dense models:}} Our work does not compare against training MoEs from scratch, a more FLOP efficient architecture in the pretraining setting \citep{fedus2022switchtransformersscalingtrillion,Databricks_2024,gale2022megablocksefficientsparsetraining}. We instead focus solely on the post-training setting. This is due to the availability of high-quality open-source dense models that practitioners may wish to use to improve quality.

\end{enumerate}

\section{Conclusion}

Our findings show that substantial additional training is required for an upcycled model to offset the extra inference overhead it introduces. While upcycling does improve model quality, it demands significant computational resources in terms of training FLOPs. Our empirical benchmarks reveal a considerable performance gap between upcycled models and dense models during inference.

Future research could explore alternative upcycling methods that address inference costs, as well as focus on further enhancing the inference performance of MoE models.

\newpage

\bibliographystyle{plainnat}
\bibliography{neurips_2024}

\medskip

\small


\appendix

\section{Experimental Details}
\label{appendix:experimental}

\subsection{Data}

For the dense pretraining phase, we use a generic common crawl data mix. For the continued pretraining/upcycling phase, we follow \cite{blakeney2024doesdatasparkjoy} and use a higher quality mix of 4 broad categories with different proportions: Large-Scale Common Crawl (15\%), code (35\%), Small-Scale Common Crawl (15\%), Domain Specific data (35\%). The choice to have a different source of data between pretraining and CPT/upcycling is meant to emulate the setting of upcycling an open-source model, where the original dataset is rarely accessible \cite{}. 

\subsection{Models}

We use the following the hyperparameters described in Table \ref{tab:430M hparams}. We performed a learning rate sweep at the shortest duration for CPT and upycling with the \textit{medium} duration dense runs, and use that learning rate for the rest of the experiments.

\subsection{Pretraining Duration}
\label{appendix:pretraining-duration}

To select our pretraining duration, we chose to focus on the ``over-trained'' setting. The seminal ``Chinchilla'' work \citep{hoffmann2022trainingcomputeoptimallargelanguage} established scaling laws for optimally pretraining LLMs for a given training FLOP budget -- finding an approximately constant optimal ratio of 20 training tokens per model parameter. In practice, models have been trained well-past this ratio of tokens/parameters, as smaller performant models are better for inference \citep{sardana2024chinchillaoptimalaccountinginferencelanguage}. For example, the Llama 3 8B and 70B models were pretrained for 15 trillion tokens, resulting in a token/parameter ratios of 1875 and 214 respectively \citep{dubey2024llama}. We include the token/parameters ratios for our model configurations in Table \ref{tab:appendix-pretraining-tokens}.

\begin{table}[h]
  \caption{Pretraining models and training durations with Token/Parameter Ratio.}
  \label{tab:appendix-pretraining-tokens}
  \centering
    \begin{tabular}{lllc}
      \toprule
      \textbf{Model Size} & \textbf{Duration} & \textbf{Tokens} & \textbf{Tokens/Parameter} \\
      \midrule
      \multirow{3}{*}{436M} & Medium      & 43.6B & 100\\
                            & Long        & 100B  & 230 \\
                            & Extra Long  & 200B  & 460 \\
      \cmidrule(r){1-4}
      \multirow{2}{*}{1.4B} & Medium      & 142B & 100 \\
                            & Long        & 354B & 250 \\
      \bottomrule
    \end{tabular}
\end{table}

\begin{table}[h]
  \centering
  \caption{430M settings}
  \label{tab:430M hparams}
  \begin{tabular}{lll}
    \toprule
    {\textbf{Model}} & 430M & 1.4B\\
    \midrule
    d\_model & 1024 & 2048 \\
    ffn hidden size & 2560 & 5120 \\
    layers & 22 & 24 \\
    heads & 16  & -- \\
    kv heads (GQA) & 4 & -- \\
    norm type & rmsnorm & -- \\

    tokenizer & gpt-4 & -- \\
    total parameters & 436M & 1.42 M\\
    \midrule
    \multicolumn{3}{l}{\textbf{Pretraining Settings}} \\
    \midrule
    optimizer & Lion & -- \\
    lr &  8e-4 & 4e-4 \\
    weight decay & 0.5 & -- \\
    batch size & 1024 & -- \\
    seq len & 4096 & -- \\
    lr scheduler & cosine & -- \\
    warmup & 1000500000 tokens & -- \\
    \midrule
    \multicolumn{2}{l}{\textbf{CPT Settings}} \\
    \midrule
    lr &  5e-5 & -- \\
    weight decay & 0.05 & --\\
    batch size & 1024 & -- \\
    seq len & 4096 & --\\
    lr scheduler & cosine & -- \\
    warmup & 5e8 tokens & -- \\
    \midrule
    \multicolumn{2}{l}{\textbf{Upcycling Settings}} \\
    \midrule
    lr &  1e-4 & 5e-5 \\
    weight decay & 0.05 & -- \\
    batch size & 1024 & -- \\
    seq len & 4096 & -- \\
    lr scheduler & cosine & -- \\
    warmup & 5e8 tokens & --\\
    load balancing loss & 0.01 & -- \\
    total parameters & 1.6B & 6.7B \\
    active parameters & 500M & 1.9B \\
    \bottomrule
  \end{tabular}
\end{table}

\subsection{CPT/Upcycling Training Duration}

We describe how much additional training was done for the corresponding models in Table \ref{tab:upcycling-and-dense-details}.

\begin{table}[h]
  \caption{CPT Model sizes and additional training amounts}
  \label{tab:upcycling-and-dense-details}
  \centering
  \begin{tabular}{llr}
    \toprule
    \textbf{Model Size} & \textbf{Duration} & \textbf{Additional Tokens} \\
    \midrule
    \multirow{5}{*}{436M}  & \multirow{5}{*}{Medium}                            & 4.3B \\
                             & & 8.7B  \\
                             & & 17.5B \\
                             & & 34.9B \\
                             & & 43.6B \\       
    \cmidrule(r){1-3}
    \multirow{5}{*}{436M}  & \multirow{5}{*}{Long}                            & 10B \\
                         & & 20B  \\
                         & & 40B \\
                         & & 80B \\
                         & & 100B \\
    \cmidrule(r){1-3}
    \multirow{5}{*}{436M}  & \multirow{5}{*}{Extra Long}                            & 20B \\
                         & & 40B  \\
                         & & 80B \\
                         & & 100B \\
                         & & 200B \\
    
   \cmidrule(r){1-3}                      
    \multirow{5}{*}{1.6B (Upcycled)} & \multirow{5}{*}{Medium}              & 3.2B \\
                             & & 6.5B  \\
                             & & 13B \\
                             & & 26B \\
                             & & 32B \\                             
    \cmidrule(r){1-3}                      
    \multirow{5}{*}{1.6B (Upcycled)} & \multirow{5}{*}{Long}                 & 7.5B \\
                             & & 15B  \\
                             & & 30B \\
                             & & 60B \\
                             & & 75B \\
    \cmidrule(r){1-3}                      
    \multirow{5}{*}{1.6B (Upcycled)} & \multirow{5}{*}{Extra Long}            & 15B \\
                             & & 30B  \\
                             & & 60B \\
                             & & 120B \\
                             & & 150B \\  
    \cmidrule(r){1-3}

    \multirow{4}{*}{1.4B}  & \multirow{4}{*}{Medium}                            & 14B \\
                             & & 28B  \\
                             & & 56B \\
                             & & 113B \\
    \cmidrule(r){1-3}
    \multirow{4}{*}{1.4B}  & \multirow{4}{*}{Long}                            & 35B \\
                         & & 70B  \\
                         & & 142B \\
                         & & 284B \\
    \cmidrule(r){1-3}
    \multirow{4}{*}{6.7B (Upcycled)}  & \multirow{4}{*}{Medium}                            & 9.6B \\
                         & & 19.3B  \\
                         & & 38.6B \\
                         & & 77.2B \\
    \cmidrule(r){1-3}                        
    \multirow{4}{*}{6.7B (Upcycled)}  & \multirow{4}{*}{Long}               & 24B \\
                         & & 48B  \\
                         & & 96B \\
                         & & 193B \\                        
    \bottomrule
  \end{tabular}
\end{table}

\subsection{Evaluation Details}
\label{appendix:eval}
We report both the smoothed final cross entropy loss and downstream Evaluation Gauntlet scores for each training run. Our reported cross entropy loss is averaged over the final 50 steps of training to account for minor fluctuations.

We evaluate our model against version 3 of the open source Evaluation Gauntlet \citep{MosaicML2023LLMEvaluation}, with tasks in five categories:
\begin{itemize}
\item \textbf{Commonsense Reasoning}:  BIG-bench Strategy QA, BIG-bench Strange Stories  \citep{srivastava2022beyond}, Common Sense QA \citep{talmor2018commonsenseqa}, COPA \citep{copa}, OpenBook QA \citep{openbook_qa}, PIQA \citep{piqa}, and SIQA \citep{sap2019socialiqa}.
\item \textbf{Language Understanding}: LAMBADA \citep{paperno2016lambada}, HellaSwag \citep{zellers2019hellaswag}, Winograd Schema Challenge \citep{winograd}, and Winogrande \citep{winogrande}.
\item \textbf{Reading Comprehension}: AGI Eval \citep{zhong2023agieval}, BoolQ \citep{clark2019boolq}, CoQA \citep{reddy-etal-2019-coqa}, and SQuAD \citep{squad}.
\item \textbf{Symbolic Problem Solving}: AGI Eval SAT Math, AGI Eval LSAT \citep{zhong2023agieval}, BIG-bench Dyck Languages \citep{srivastava2022beyond}, BIG-bench Elementary Math QA \citep{srivastava2022beyond}, BIG-bench Operators \citep{srivastava2022beyond}, GSM8k \citep{cobbe2021gsm8k}, LogiQA \citep{logiqa}, Math QA \citep{math_qa}, and SVAMP \citep{patel-etal-2021-nlp}.
\item \textbf{World Knowledge}:  ARC Easy, ARC Challenge \citep{arc}, BIG-bench WikiData \citep{srivastava2022beyond}, Jeopardy \citep{jeopardy}, and MMLU \citep{hendrycks2020measuring}.
\end{itemize}

The Gauntlet Core Average is a simple average accuracy over all tasks. Each task is equally weighted after subtracting out the task's baseline random accuracy and normalizing.

\section{Inference Details}\label{appendix:inference}

We use the vLLM v0.6.2 \citep{kwon2023efficient} inference engine for all inference benchmarks. For fair comparisons across MoE and dense vLLM implementations, we compare the performance of the Mistral and Mixtral architectures \citep{jiang2024mixtralexperts}.
We use the tiktoken tokenizer to be consistent with the training setup. We vary the Request Per Seconds at 0.1 increments in order to obtain the curves for Latency vs Throughput.

For the 430M and 1.4B models and their upcycled counterparts, we run inference on a single H100 GPU. For the 8B model and its upcycled counterpart we use 4 H100 GPUs and use tensor parallelism (TP). Note that we could have fit the 8B model \textit{without} TP on a single H100, but chose to use the same number of GPUs for both the dense and upcycled versions.

We use 3500 input tokens and 300 output tokens. We measure latency (the total time in seconds to process input + output tokens per request) and throughput (the amount of input + output tokens that the system can process per second). We average our results over 5 runs.

\begin{figure}[h]
    \centering
        \includegraphics[height=4cm, trim={0cm 0cm 0cm 0cm}, clip]{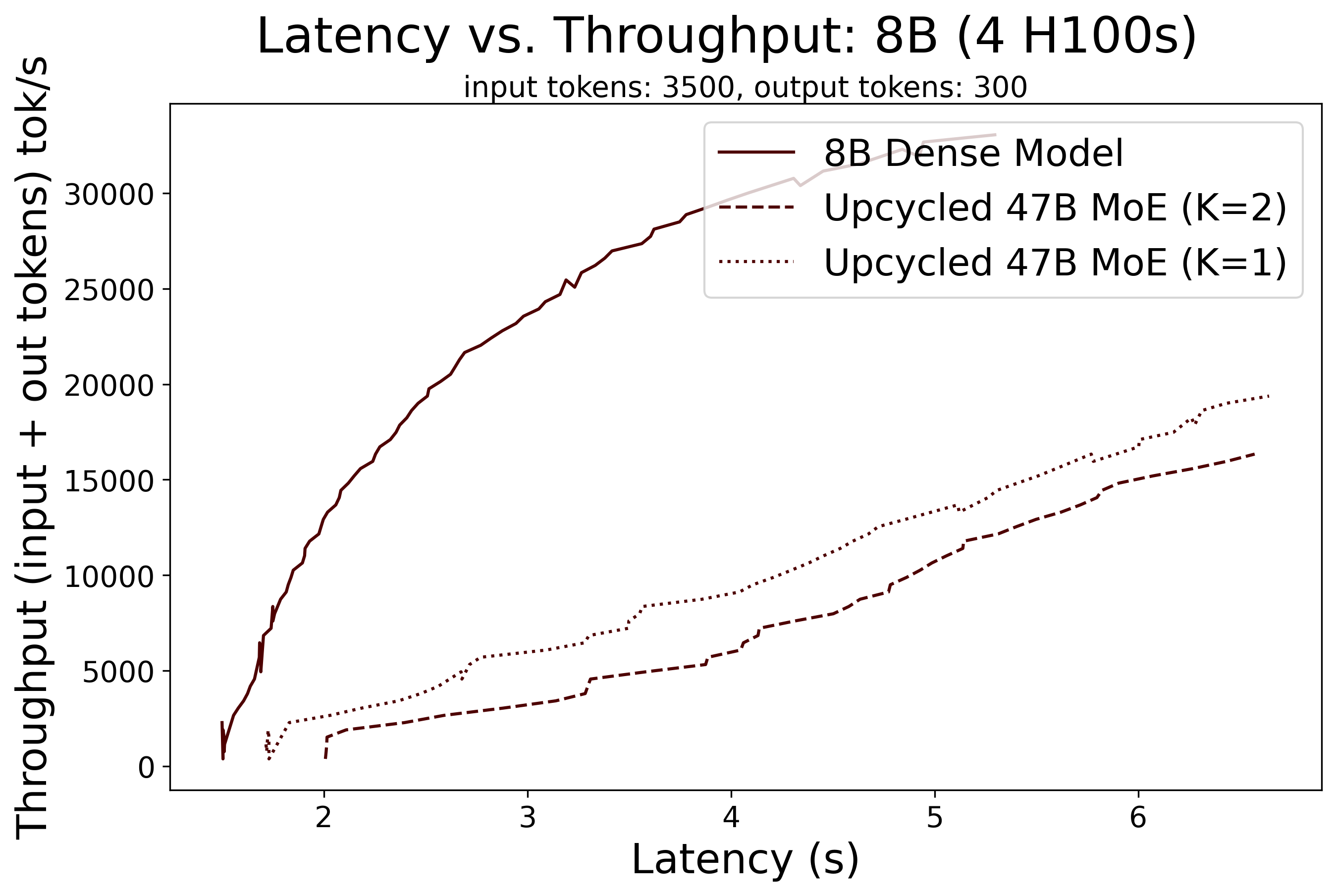}
    \caption{Latency vs. Throughput - 8B vs 47B MoE}
\end{figure}

\section{Additional Results}
\begin{figure}[h]
    \centering
    \begin{subfigure}{0.45\textwidth}
        \centering
        \includegraphics[width=\linewidth]{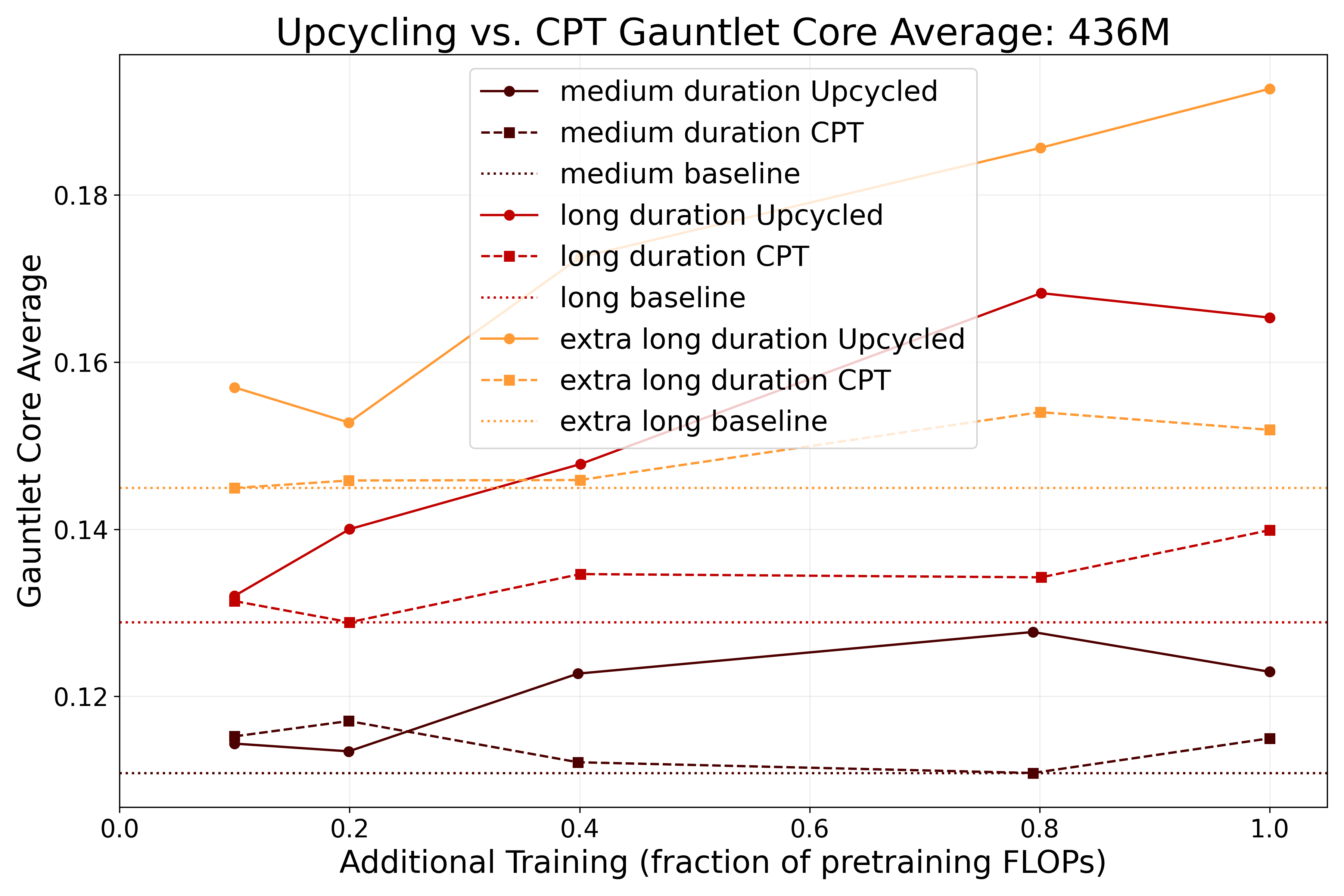}
    \end{subfigure}
    \hfill
    \begin{subfigure}{0.45\textwidth}
        \centering
        \includegraphics[width=\linewidth]{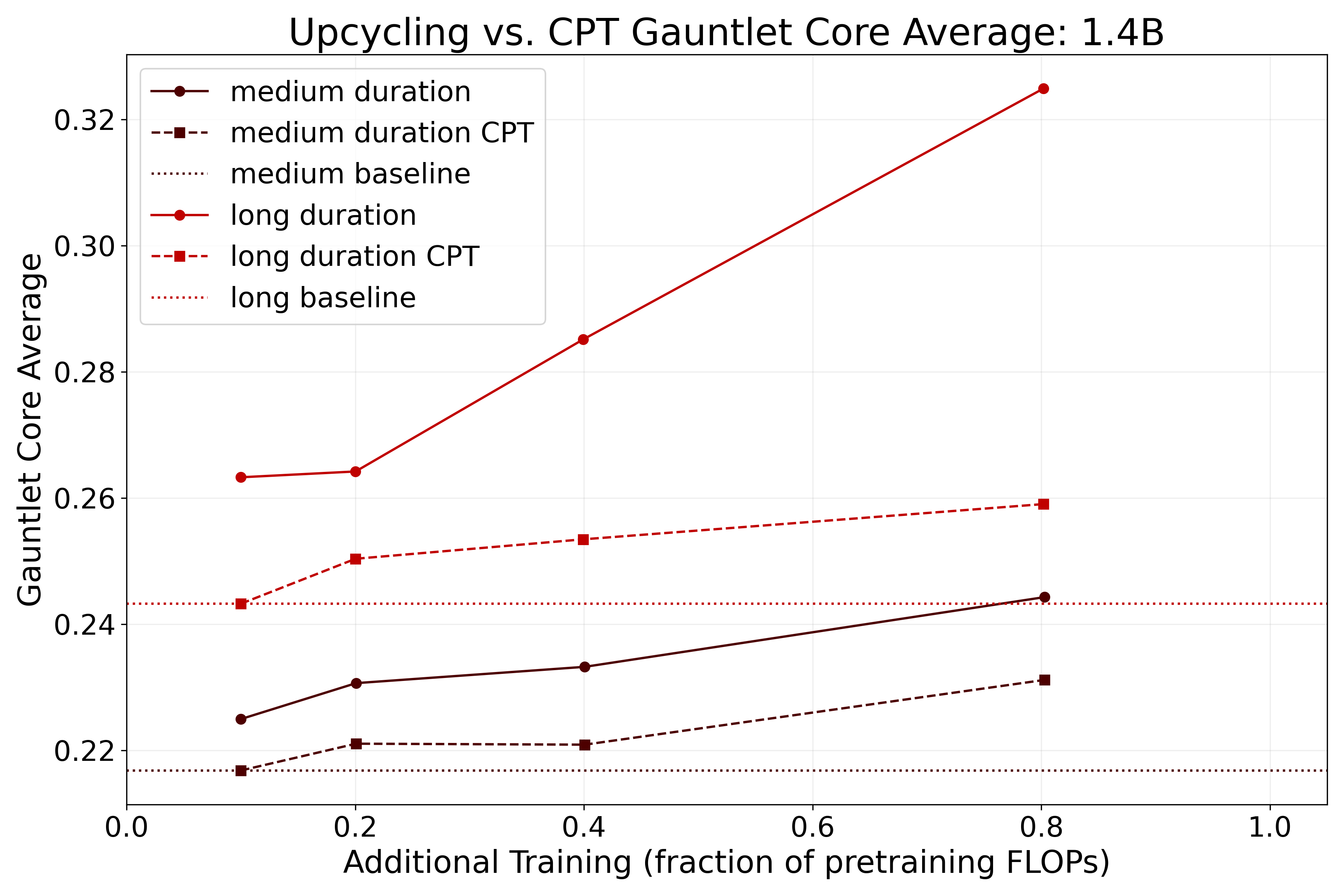}
    \end{subfigure}
    \caption{Absolute Gauntlet Scores for Dense vs. Upcycled Models.} 
    \vspace{-5mm}
    \label{fig:raw-gauntlet}
\end{figure}

We show plots with absolute Gauntlet scores in Figure \ref{fig:raw-gauntlet}. We find that CPT models seem to saturate dramatically as compared to the CE loss shown in Figure \ref{fig:main-train-results}, which is an interesting disconnect that requires more investigation.

\end{document}